\pdfoutput=1

\documentclass[conference]{IEEEtran}

\pdfoutput=1
\IEEEoverridecommandlockouts
\usepackage{cite}
\usepackage{amsmath,amssymb,amsfonts}
\usepackage{algorithmic}
\usepackage{graphicx}
\usepackage{textcomp}
\usepackage{xcolor}
\usepackage{hyperref}

\usepackage[normalem]{ulem}

\def\BibTeX{{\rm B\kern-.05em{\sc i\kern-.025em b}\kern-.08em
    T\kern-.1667em\lower.7ex\hbox{E}\kern-.125emX}}

\usepackage{fancyhdr}

\fancypagestyle{firstpage}{%
  \fancyhf{} 
  \fancyhead[L]{\small This paper has been accepted at 2025 International Conference on Intelligent Communication Networks and Computational Techniques (ICICNCT) 
  }

  \fancyfoot[L]{\small 979-8-3315-8623-2/25/\$31.00 \textcopyright~2025 IEEE}

}

\begin{document}

\title{Enhancing Polyp Segmentation via Encoder Attention and Dynamic Kernel Update\\
}

\author{
\IEEEauthorblockN{Fatemeh Salahi Chashmi\textsuperscript{*}, Roya Sotoudeh\textsuperscript{*}} 

\IEEEauthorblockA{
\textit{Faculty of New Sciences and Technologies, University of Tehran, Tehran, Iran} \\
fatemeh.salahi@ut.ac.ir
}

\IEEEauthorblockA{
\textit{Department of Electrical Engineering, Urmia University, Urmia, Iran} \\
r.sotudeh@urmia.ac.ir
}

\thanks{\textsuperscript{*}These authors contributed equally to this work.}
}

\maketitle
\thispagestyle{firstpage} 

\begin{abstract} 
Polyp segmentation is a critical step in colorectal
cancer detection, but it remains challenging due to the varied
shapes, sizes, and low-contrast boundaries of polyps in medical
imaging. In this work, we introduce a novel framework designed
to improve segmentation accuracy and efficiency by incorporat-
ing a Dynamic Kernel (DK) mechanism and a global Encoder
Attention (EA) module. The DK mechanism, initialized by a
global context vector from the EA module, iteratively refines
segmentation predictions across decoding stages, allowing the
model to focus on and accurately delineate complex polyp bound-
aries. The EA module enhances the network’s ability to capture
critical lesion features by aggregating multi-scale features from
all encoder layers. Additionally, we implement Unified Channel
Adaptation (UCA) in the decoder, which standardizes feature
dimensions across stages, ensuring consistent and computation-
ally efficient information fusion. We clarify that our approach
builds upon the lesion-aware kernel framework [1] by introducing
a more flexible, attention-driven kernel initialization and a
unified decoder design. Extensive experiments on the Kvasir-SEG
and CVC-ClinicDB benchmark datasets demonstrate that our
model outperforms several state-of-the-art segmentation models,
achieving superior Dice and Intersection over Union (IoU)
scores. The UCA simplifies the decoder architecture, resulting
in reduced computational cost without compromising accuracy.
Our proposed method thus provides a robust, adaptable solution
for polyp segmentation, with potential applications in clinical and
automated diagnostic systems.
\end{abstract}

\section{Introduction}

Colorectal cancer (CRC) is one of the most common tumors globally, often originating from polyps in the adenomatous tissue \cite{siegel2018cancer}. {These} polyps, if not treated promptly and appropriately, have a heightened risk of transforming into malignant tumors. Colonoscopy is the definitive method for detecting colorectal lesions, as it enables the timely identification and excision of colorectal polyps, thereby preventing their proliferation. The objective of polyp segmentation is to precisely identify polyps in their first phases, a critical issue in medical image processing essential for the clinical prevention of rectal cancer. Nevertheless, the varied and often concealed characteristics of polyps render precise polyp segmentation challenging. Traditional approaches employed hand-crafted features for polyp identification; however, they struggled with complex circumstances and exhibited a high misdiagnosis rate \cite{karkehabadi2024evaluating}\cite{mamonov2014automated}\cite{tajbakhsh2015automated}.
 With the advancement of deep learning, polyp segmentation has made encouraging steps to further improve  diagnostics. One main approach to localize  polyps  is  {through the use of Artificial Intelligence (AI) models}. These models,  {known as segmentation models, are} computer vision approaches that  {divide} the input image into distinct groupings of pixels, or classes.  {The aim of medical image segmentation is to provide a precise representation of items of interest within an image for diagnosis, treatment planning, and quantitative analysis.}
Among the introduced AI models for segmentation, the U-shaped method called UNet \cite{ronneberger2015u}, particularly has attracted a lot of interest because of its capacity to incorporate multi-level characteristics for high-resolution result reconstruction. UNet employs an encoder-decoder architecture including skip connections, facilitating the integration of multi-level information utilized in successive upsampling operations to generate a high-resolution segmentation map.

{While dynamic kernels have shown promise, their effectiveness hinges on the quality of the initial context representation. This study introduces an innovative segmentation framework designed to address this limitation. Our primary contribution is a novel integration of a global Encoder Attention (EA) module with a Dynamic Kernel (DK) mechanism, which allows the model to learn a more robust initial kernel by aggregating features from all encoder stages. This contrasts with previous works like LDNet \cite{zhang2022lesion}, which derive context from only the deepest encoder layer. Our second contribution is a Unified Channel Adaptation (UCA) strategy in the decoder, which standardizes feature dimensions to a constant C=32. This simplifies the decoder architecture, reduces parameters, and ensures efficient feature fusion without the need for complex, multi-scale channel adjustments seen in other models. Collectively, our EA-driven kernel initialization and streamlined UCA decoder create a powerful, efficient, and generalizable model for polyp segmentation, as validated by extensive experiments.

\section{Material and Method}
\subsection{General Overview}
We propose our model for polyp segmentation, leveraging a novel Encoder Attention (EA) module to enhance feature representation while keeping the general architecture simple. Our model is an extension of the lesion-aware segmentation model introduced in \cite{zhang2022lesion}, improving upon it with a new global context extraction method and a streamlined decoder. The primary components of our model are: First, a dynamic kernel generation and update scheme, second, an encoder attention mechanism, and third, an efficient decoder with unified channels. Figure~\ref{fig:architecture} shows the complete pipeline and the architecture.

\begin{figure*}[t!]
  \centering
  \includegraphics[width=0.8\textwidth]{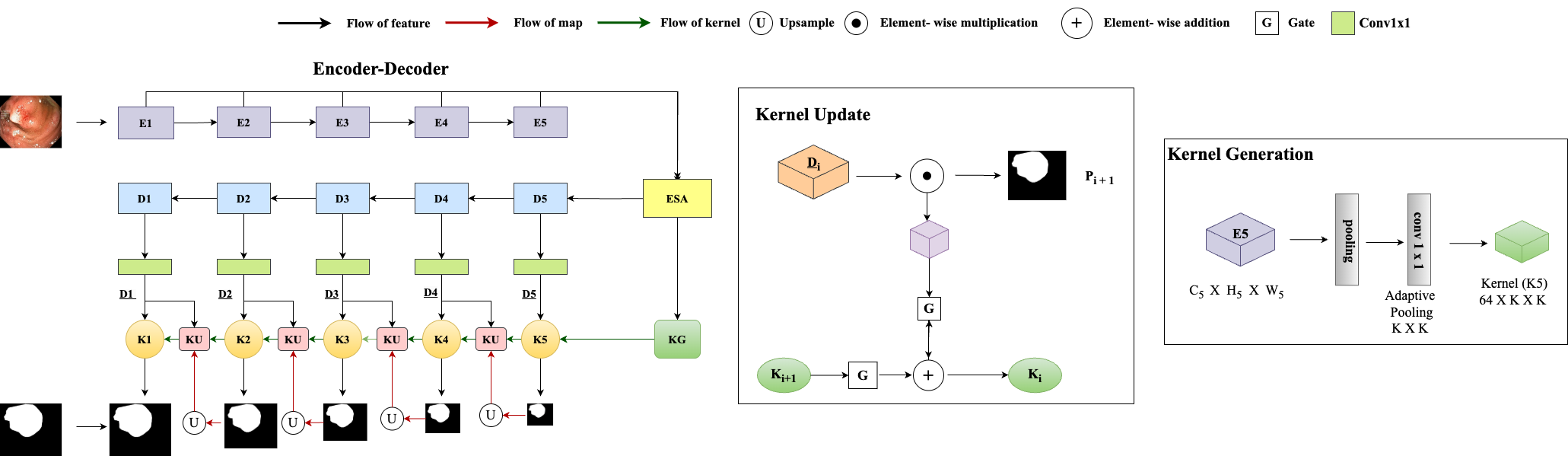}
  \caption[Overall architecture of the proposed model]{%
    The overall architecture of our proposed model. 
    \textbf{Left (Encoder):} A Res2Net-50 backbone extracts multi-scale features
    ($F_1$--$F_5$). 
    \textbf{Top (EA Module):} The Encoder Attention (EA) module takes features from all encoder
    stages, applies global average pooling, and uses self-attention to generate a single global
    context vector $\mathbf{G}$. 
    \textbf{Center (Dynamic Kernel):} The context vector $\mathbf{G}$ is transformed into an
    initial dynamic kernel $\mathbf{K}_5$. This kernel is iteratively updated at each decoder stage
    ($i = 4, 3, 2, 1$) using features from the decoder and the prediction from the previous stage
    ($\mathbf{P}_{i+1}$) to produce a refined kernel $\mathbf{K}_i$ and a new prediction
    $\mathbf{P}_i$. 
    \textbf{Right (Decoder):} The decoder employs Unified Channel Adaptation (UCA) to standardize
    all incoming features to a fixed channel dimension ($C = 32$) before fusion and upsampling,
    finally producing the segmentation map.%
  }
  \label{fig:architecture}
\end{figure*}
\subsection{Encoder with Global Encoder Attention Mechanism}
Our encoder is based on the Res2Net\cite{gao2019res2net} backbone, which extracts multi-scale features from the input image across five stages. In contrast to conventional segmentation models that rely solely on the deepest layer for global feature extraction, we introduce an Encoder Attention (EA) module to capture the most relevant features across all encoder stages. Specifically, the EA module is designed to adaptively attend to features from each encoder layer by computing query, key, and value representations.
{Given the output feature map from each encoder stage, denoted as \( F_i \in \mathbb{R}^{C_i \times H_i \times W_i} \) for \( i = 1, \ldots, 5 \), the EA module first applies global average pooling to obtain a compact vector representation \( P_i \in \mathbb{R}^{C_i} \). These pooled vectors are then used to compute attention scores via a standard self-attention mechanism:}
\[
\operatorname{Attention}(Q, K, V) = \operatorname{softmax}\left(\frac{QK^T}{\sqrt{d_k}}\right)V,
\]

{where \( Q \), \( K \), and \( V \) are computed from the pooled feature vectors. The resulting attended feature map generated by the EA module serves as an enhanced global context representation \( G \in \mathbb{R}^{C_{\text{out}}} \), which is then used to initialize the dynamic segmentation kernel.}

\subsection{Dynamic Kernel for Segmentation Head}

The dynamic kernel mechanism forms the core of the segmentation head , allowing for an adaptive and iterative refinement of segmentation predictions across decoder stages. This mechanism is initialized using {the} global context vector  {G generated by the EA module}, which aggregates multi-scale features from each encoder stage to capture the most relevant contextual information. By conditioning the initial kernel parameters on this enhanced global context, the segmentation head is better equipped to generalize across diverse polyp shapes, textures, and sizes.
\subsubsection{Initial Kernel Generation}

Let the global context vector be denoted as \( G \in \mathbb{R}^{d_k} \). 

{This vector is passed through a small multi-layer perceptron (MLP), denoted as \( \phi \), to generate the initial parameters for the dynamic kernel \( K_5 \in \mathbb{R}^{C_d \times 1 \times 1} \), where \( C_d \) is the unified channel dimension of the decoder.}
\begin{equation}
\label{eq:initial_kernel}
K_5 = \phi(G)
\end{equation}

As the decoding process proceeds, from stage \(i = 4\) down to \(1\), the kernel is iteratively updated, leveraging both the features from the current decoder layer and the segmentation predictions from the previous stage. This iterative process is designed to gradually refine the segmentation boundaries by enabling the kernel to incorporate and respond to local feature variations at each stage.

At each stage \(i\), let \(D_i \in \mathbb{R}^{C_d \times H_i \times W_i}\) denote the unified feature map from the decoder, and let \(P_{i+1} \in \mathbb{R}^{1 \times H_{i+1} \times W_{i+1}}\) represent the segmentation prediction from the previous stage.  {\(P_{i+1}\) is first upsampled to match the spatial dimensions of \(D_i\). The kernel update process begins by generating a lesion-aware feature aggregate \(A_i \in \mathbb{R}^{C_d}\):}
\begin{align}
\label{eq:assembly}
&{\sout{A_i = \sum_{h=1}^{H_i} \sum_{w=1}^{W_i} F_i(h, w) \cdot \text{upsample}(P_{i+1})(h, w),}} \nonumber \\
&{A_i = \sum_{h,w} D_i(h, w) \odot \text{upsample}(P_{i+1})(h, w),}
\end{align}
The element-wise multiplication operation} ensures that the {aggregation} focuses on regions already identified as probable lesion areas.

\subsubsection{Feature Splitting and Gate Mechanism}

The {aggregated feature} ($A_i$) is then linearly transformed

\begin{equation}
\label{eq:split}
{[A_i^{\text{feat}}, A_i^{\text{gate}}] = \text{Linear}(A_i).}
\end{equation}

{To refine the kernel update, we employ a gating mechanism that modulates the influence of the aggregated features. The gate value $g_i$ is computed as:}

\begin{equation}
\label{eq:gate}
 \\
{g_i = \sigma(W_g [A_i^{\text{gate}} \odot K_{i+1}] + b_g),}
\end{equation}

where $\sigma$ denotes the sigmoid function, 
{$W_g$ and $b_g$ are learnable parameters of a linear layer, and $K_{i+1}$ is the kernel from the previous stage.} 
This gating operation selectively combines information from the previous prediction with current decoder features, mitigating noise and focusing on the most discriminative information.

\subsubsection{Kernel Update and Final Prediction}

The refined kernel ($K_i$) for the current stage is computed by integrating the gated  {feature:}
\begin{align}
\label{eq:update}
{\sout{K_i = G_i \cdot A_i^{\text{out}} + (1 - G_i) \cdot K_{i+1}^{\text{out}}.}} \nonumber \\
{K_i = g_i \odot A_i^{\text{feat}} + (1 - g_i) \odot K_{i+1}.}
\end{align}
This updated kernel is then convolved with the current decoder features ($D_i$) to generate a new segmentation prediction ($P_i$).

\subsection{Decoder with Unified Channel Adaptation}
The decoder in our architecture is engineered to process and enhance features derived from the encoder via a series of upsampling and feature transformation phases.
In conventional segmentation architectures, variations in feature dimensions between different encoder and decoder layers can lead to inconsistencies and hinder effective feature fusion. To address this, we introduce {our Unified Channel Adaptation (UCA) strategy. At each decoder stage i, the incoming feature map from the corresponding encoder stage is processed by a 1x1 convolution to reduce its channel dimension to a fixed, small number (C 
d

 =32). This unification procedure synchronizes feature representations throughout the network, enabling more seamless transitions between stages and enhancing feature aggregation reliability.

{After channel reduction, each stage in the decoder upsamples the feature map from the previous stage and fuses it with the channel-adapted encoder feature. The iterative kernel update mechanism described in Equations \ref{eq:assembly}-\ref{eq:update} is then applied to these fused features to refine the segmentation map.}
The unified channel adaptation not only streamlines feature fusion but also markedly decreases computing complexity. By maintaining a uniform feature dimension across the decoder, the model's efficiency is enhanced while retaining segmentation performance. This optimized method enables the decoder to {maintain} high precision across all scales while minimizing resource requirements, making it suitable for clinical environments.

\section{Implementation}

{We trained our model end-to-end using} the \emph{Res2Net-50} backbone initialized with ImageNet weights, allowing the model to {leverage pre-trained features while adapting to the polyp segmentation task}. The model was trained using a combination of Binary Cross-Entropy (\( \mathcal{L}_{\text{BCE}} \)) and Dice loss (\( \mathcal{L}_{\text{Dice}} \)),  to optimize both pixel-wise classification accuracy and segmentation region overlap. The total loss (\( \mathcal{L}_{\text{total}} \)) is defined as:

\begin{equation}
\label{eq:loss}
{\mathcal{L}_{\text{total}} = \mathcal{L}_{\text{BCE}} + \mathcal{L}_{\text{Dice}}.}
\end{equation}

We used the Stochastic Gradient Descent (SGD) optimizer with a momentum of 0.9 and a weight decay of \( 10^{-5} \). The initial learning rate was set to {4e\text{-}4} and decayed according to a polynomial learning rate schedule.  
Data augmentations including random horizontal and vertical flips, rotations, and cropping---were applied to improve model generalization. The batch size was set to 32, and all images were resized to \( 256 \times 256 \) to ensure computational efficiency and consistency.

\section{Experiments and Results}
We evaluated our method on two public polyp segmentation benchmark datasets: \textbf{Kvasir-SEG} \cite{jha2020kvasir} and \textbf{CVC-ClinicDB} \cite{bernal2015wm}. To ensure consistency in evaluation, we followed the standard data split from prior works, using 80

\subsection{Evaluation Metrics}

We assessed the segmentation performance using the following standard metrics: recall, specificity, precision, Dice score, intersection over union for polyp (IoU$_p$), IoU for background (IoU$_b$), mean IoU (mIoU), and accuracy.

Recall measures the proportion of actual polyp pixels that were correctly identified, while specificity quantifies how well the model avoids false positives in background regions. Precision reflects the proportion of predicted polyp pixels that are truly polyp pixels.

The Dice score, a widely used metric in medical image segmentation, measures the overlap between the predicted segmentation and the ground truth, making it especially useful for evaluating how well the model captures complex polyp boundaries.

IoU$_p$ and IoU$_b$ compute the intersection-over-union for the polyp and background classes, respectively, offering a class-wise perspective on segmentation performance. Mean IoU (mIoU) is the average of IoU$_p$ and IoU$_b$, providing a balanced overall measure of segmentation quality. Finally, accuracy indicates the proportion of correctly classified pixels across the entire image.
\subsection{Results and Analysis}

\begin{table*}[ht!]
\centering
\caption{Comparison with other state-of-the-art methods on two benchmark datasets.}
\begin{tabular}{l|cccccccc}
\hline
\textbf{Method} & \textbf{Recall} & \textbf{Spec} & \textbf{Prec} & \textbf{Dice} & \textbf{IoU$_{p}$} & \textbf{IoU$_{b}$} & \textbf{mIoU} & \textbf{Acc} \\ \hline
\multicolumn{9}{c}{\textbf{Kvasir}} \\ \hline
UNet \cite{ronneberger2015u} & 87.04 & 97.25 & 84.28 & 82.60 & 73.39 & 93.89 & 83.64 & 95.05 \\
ResUNet \cite{zhang2018road} & 84.70 & 97.17 & 83.00 & 80.50 & 70.60 & 93.19 & 81.89 & 94.43 \\
UNet++ \cite{zhou2018unet++} & 89.23 & 97.20 & 85.57 & 84.77 & 76.42 & 94.23 & 85.32 & 95.44 \\
ACSNet \cite{zhang2020adaptive} & 91.35 & 98.39 & 91.46 & 89.54 & 83.72 & 96.42 & 90.07 & 97.16 \\
PraNet \cite{fan2020pranet} & 93.90 & 97.33 & 89.87 & 90.32 & 84.55 & 95.98 & 90.26 & 96.75 \\
CCBANet \cite{nguyen2021ccbanet} & 90.71 & 98.04 & 91.02 & 89.04 & 82.82 & 96.21 & 89.52 & 97.02 \\
SANet \cite{wei2021shallow} & 92.06 & 98.20 & 91.14 & 89.92 & 83.97 & 96.54 & 90.26 & 97.18 \\
{LDNet } \cite{zhang2022lesion} & 92.72 & 98.05 & 92.04 & 90.70 & 85.30 & 96.71 & 91.01 & 97.35 \\ \hline
\textbf{Ours} & 94.11 & 98.68 & 93.24 & 93.74 & 88.16 & 97.71 & 92.93 & 98.31 \\ \hline

\multicolumn{9}{c}{\textbf{CVC-ClinicDB}} \\ \hline
UNet \cite{ronneberger2015u} & 88.61 & 98.70 & 85.10 & 85.12 & 77.78 & 97.70 & 87.74 & 97.95 \\
ResUNet \cite{zhang2018road} & 90.89 & 99.25 & 90.22 & 89.98 & 82.77 & 98.18 & 90.47 & 98.37 \\
UNet++ \cite{zhou2018unet++} & 87.78 & 99.21 & 90.02 & 87.99 & 80.69 & 97.92 & 89.30 & 98.12 \\
ACSNet \cite{zhang2020adaptive} & 93.46 & 99.54 & 94.63 & 93.80 & 88.57 & 98.95 & 93.76 & 99.08 \\
PraNet \cite{fan2020pranet} & 95.22 & 99.34 & 92.25 & 93.49 & 88.08 & 98.92 & 93.50 & 99.05 \\
CCBANet \cite{nguyen2021ccbanet} & 94.89 & 99.22 & 91.39 & 92.83 & 86.96 & 98.79 & 92.87 & 98.93 \\
SANet \cite{wei2021shallow} & 94.74 & 99.41 & 92.88 & 93.61 & 88.26 & 98.94 & 93.60 & 99.07 \\
{LDNet} \cite{zhang2022lesion} & 94.49 & 99.51 & 94.53 & 94.31 & 89.48 & 98.95 & 94.21 & 99.08 \\
\textbf{Ours} & 95.16 & 99.14 & 94.50 & 94.89 & 90.29 & 98.50 & 94.40 & 98.88 \\ \hline

\end{tabular}
\label{tab:comparison}
\end{table*}

Table~\ref{tab:comparison} presents the quantitative results of our method compared to state-of-the-art segmentation models, including UNet \cite{ronneberger2015u}, ResUNet \cite{zhang2018road}, UNet++ \cite{zhou2018unet++}, ACSNet \cite{zhang2020adaptive}, PraNet \cite{fan2020pranet}, CCBANet \cite{nguyen2021ccbanet}, and SANet \cite{wei2021shallow}. Across four benchmark datasets, it consistently achieved high scores on key metrics such as Dice Score, Intersection over Union (IoU), Precision, and Accuracy, underscoring its effectiveness in polyp segmentation.

\subsection{Performance on Kvasir-SEG Dataset}

On the Kvasir-SEG dataset, our method achieved a Dice score of 92.93\% and an IoU for polyps (IoU$_p$) of 88.16\%, outperforming other models in the majority of metrics. A recall of 94.11\% and precision of 93.24\% demonstrate its ability to accurately identify polyp regions with minimal false positives, showing a balanced trade-off between sensitivity and specificity. The high IoU$_b$ score of 97.71\% for the background region further illustrates the model's ability to accurately distinguish polyp boundaries from surrounding tissues, which is crucial in medical applications where precision is paramount.

\subsection{Performance on CVC-ClinicDB Dataset}

On the CVC-ClinicDB dataset, our method also showed superior performance, achieving a Dice score of 94.40\%, with an IoU$_p$ of 90.29\% and mIoU of 94.40\%. These results highlight our method's robust generalization capability on seen datasets, where it effectively segments polyps of varying sizes and shapes. Compared to other methods, it also obtained the highest specificity (99.14\%) and precision (94.50\%), indicating a strong ability to minimize false positive detections. These results are essential for clinical applicability, as reducing false positives can help avoid unnecessary medical procedures.

\section{Conclusion}

We presented a novel framework for polyp segmentation that leverages a Dynamic Kernel mechanism and EA module. It was designed to address the inherent challenges in polyp segmentation, including diverse shapes, sizes, and low-contrast boundaries, by dynamically adapting its segmentation kernel based on a global context derived from encoder features. Our model also integrates Unified Channel Adaptation in the decoder, which ensures consistent feature dimensions across stages and improves computational efficiency.
Through extensive experiments on two benchmark datasets including Kvasir-SEG and CVC-ClinicDB, our method demonstrated superior performance over several state-of-the-art segmentation models, achieving higher Dice and IoU scores across challenging scenarios. The proposed framework outperforms the existing methods. By accurately delineating polyp boundaries, it has the potential to aid early colorectal cancer detection, providing a robust and reliable tool for endoscopic image analysis.

\bibliographystyle{IEEEtran}
\bibliography{ref}

\begin{thebibliography}{10}
\providecommand{\url}[1]{#1}
\csname url@samestyle\endcsname
\providecommand{\newblock}{\relax}
\providecommand{\bibinfo}[2]{#2}
\providecommand{\BIBentrySTDinterwordspacing}{\spaceskip=0pt\relax}
\providecommand{\BIBentryALTinterwordstretchfactor}{4}
\providecommand{\BIBentryALTinterwordspacing}{\spaceskip=\fontdimen2\font plus
\BIBentryALTinterwordstretchfactor\fontdimen3\font minus \fontdimen4\font\relax}
\providecommand{\BIBforeignlanguage}[2]{{%
\expandafter\ifx\csname l@#1\endcsname\relax
\typeout{** WARNING: IEEEtran.bst: No hyphenation pattern has been}%
\typeout{** loaded for the language `#1'. Using the pattern for}%
\typeout{** the default language instead.}%
\else
\language=\csname l@#1\endcsname
\fi
#2}}
\providecommand{\BIBdecl}{\relax}
\BIBdecl

\bibitem{siegel2018cancer}
R.~L. Siegel, K.~D. Miller, and A.~Jemal, ``Cancer statistics, 2018,'' \emph{CA: a cancer journal for clinicians}, vol.~68, no.~1, pp. 7--30, 2018.

\bibitem{karkehabadi2024evaluating}
A.~Karkehabadi and S.~Sadeghmalakabadi, ``Evaluating deep learning models for architectural image classification: A case study on the uc davis campus,'' in \emph{2024 IEEE 8th International Conference on Information and Communication Technology (CICT)}.\hskip 1em plus 0.5em minus 0.4em\relax IEEE, 2024, pp. 1--6.

\bibitem{mamonov2014automated}
A.~V. Mamonov, I.~N. Figueiredo, P.~N. Figueiredo, and Y.-H.~R. Tsai, ``Automated polyp detection in colon capsule endoscopy,'' \emph{IEEE transactions on medical imaging}, vol.~33, no.~7, pp. 1488--1502, 2014.

\bibitem{tajbakhsh2015automated}
N.~Tajbakhsh, S.~R. Gurudu, and J.~Liang, ``Automated polyp detection in colonoscopy videos using shape and context information,'' \emph{IEEE transactions on medical imaging}, vol.~35, no.~2, pp. 630--644, 2015.

\bibitem{ronneberger2015u}
O.~Ronneberger, P.~Fischer, and T.~Brox, ``U-net: Convolutional networks for biomedical image segmentation,'' in \emph{Medical image computing and computer-assisted intervention--MICCAI 2015: 18th international conference, Munich, Germany, October 5-9, 2015, proceedings, part III 18}.\hskip 1em plus 0.5em minus 0.4em\relax Springer, 2015, pp. 234--241.

\bibitem{zhang2022lesion}
R.~Zhang, P.~Lai, X.~Wan, D.-J. Fan, F.~Gao, X.-J. Wu, and G.~Li, ``Lesion-aware dynamic kernel for polyp segmentation,'' in \emph{International Conference on Medical Image Computing and Computer-Assisted Intervention}.\hskip 1em plus 0.5em minus 0.4em\relax Springer, 2022, pp. 99--109.

\bibitem{gao2019res2net}
S.-H. Gao, M.-M. Cheng, K.~Zhao, X.-Y. Zhang, M.-H. Yang, and P.~Torr, ``Res2net: A new multi-scale backbone architecture,'' \emph{IEEE transactions on pattern analysis and machine intelligence}, vol.~43, no.~2, pp. 652--662, 2019.

\bibitem{jha2020kvasir}
D.~Jha, P.~H. Smedsrud, M.~A. Riegler, P.~Halvorsen, T.~De~Lange, D.~Johansen, and H.~D. Johansen, ``Kvasir-seg: A segmented polyp dataset,'' in \emph{MultiMedia modeling: 26th international conference, MMM 2020, Daejeon, South Korea, January 5--8, 2020, proceedings, part II 26}.\hskip 1em plus 0.5em minus 0.4em\relax Springer, 2020, pp. 451--462.

\bibitem{bernal2015wm}
J.~Bernal, F.~J. S{\'a}nchez, G.~Fern{\'a}ndez-Esparrach, D.~Gil, C.~Rodr{\'\i}guez, and F.~Vilari{\~n}o, ``Wm-dova maps for accurate polyp highlighting in colonoscopy: Validation vs. saliency maps from physicians,'' \emph{Computerized medical imaging and graphics}, vol.~43, pp. 99--111, 2015.

\bibitem{zhang2018road}
Z.~Zhang, Q.~Liu, and Y.~Wang, ``Road extraction by deep residual u-net,'' \emph{IEEE Geoscience and Remote Sensing Letters}, vol.~15, no.~5, pp. 749--753, 2018.

\bibitem{zhou2018unet++}
Z.~Zhou, M.~M. Rahman~Siddiquee, N.~Tajbakhsh, and J.~Liang, ``Unet++: A nested u-net architecture for medical image segmentation,'' in \emph{Deep Learning in Medical Image Analysis and Multimodal Learning for Clinical Decision Support: 4th International Workshop, DLMIA 2018, and 8th International Workshop, ML-CDS 2018, Held in Conjunction with MICCAI 2018, Granada, Spain, September 20, 2018, Proceedings 4}.\hskip 1em plus 0.5em minus 0.4em\relax Springer, 2018, pp. 3--11.

\bibitem{zhang2020adaptive}
R.~Zhang, G.~Li, Z.~Li, S.~Cui, D.~Qian, and Y.~Yu, ``Adaptive context selection for polyp segmentation,'' in \emph{Medical Image Computing and Computer Assisted Intervention--MICCAI 2020: 23rd International Conference, Lima, Peru, October 4--8, 2020, Proceedings, Part VI 23}.\hskip 1em plus 0.5em minus 0.4em\relax Springer, 2020, pp. 253--262.

\bibitem{fan2020pranet}
D.-P. Fan, G.-P. Ji, T.~Zhou, G.~Chen, H.~Fu, J.~Shen, and L.~Shao, ``Pranet: Parallel reverse attention network for polyp segmentation,'' in \emph{International conference on medical image computing and computer-assisted intervention}.\hskip 1em plus 0.5em minus 0.4em\relax Springer, 2020, pp. 263--273.

\bibitem{nguyen2021ccbanet}
T.-C. Nguyen, T.-P. Nguyen, G.-H. Diep, A.-H. Tran-Dinh, T.~V. Nguyen, and M.-T. Tran, ``Ccbanet: cascading context and balancing attention for polyp segmentation,'' in \emph{Medical Image Computing and Computer Assisted Intervention--MICCAI 2021: 24th International Conference, Strasbourg, France, September 27--October 1, 2021, Proceedings, Part I 24}.\hskip 1em plus 0.5em minus 0.4em\relax Springer, 2021, pp. 633--643.

\bibitem{wei2021shallow}
J.~Wei, Y.~Hu, R.~Zhang, Z.~Li, S.~K. Zhou, and S.~Cui, ``Shallow attention network for polyp segmentation,'' in \emph{Medical Image Computing and Computer Assisted Intervention--MICCAI 2021: 24th International Conference, Strasbourg, France, September 27--October 1, 2021, Proceedings, Part I 24}.\hskip 1em plus 0.5em minus 0.4em\relax Springer, 2021, pp. 699--708.

\end{thebibliography}
\end{document}